\documentclass[runningheads]{llncs}
\usepackage[T1]{fontenc}
\usepackage{graphicx}
\usepackage{float}
\usepackage{caption}
\usepackage{subcaption}
\usepackage{adjustbox}
\usepackage{threeparttable}
\usepackage[backref=page]{hyperref}
\usepackage{orcidlink}
\usepackage{xcolor}
\hypersetup{
    colorlinks=true,
    linkcolor=blue,
    urlcolor=blue,
    citecolor=black,
    filecolor=black,
    linkbordercolor=white, 
    urlbordercolor=white,
    citebordercolor=white,
    filebordercolor=white,
}

\usepackage{hhline}
\usepackage{multirow}
\usepackage{makecell}
\usepackage{comment}

\usepackage{siunitx}

\usepackage{float} 
\usepackage[noabbrev]{cleveref}

\usepackage{color}

\usepackage[misc]{ifsym}

\begin{document}
\title{\texorpdfstring{A Multi-Stage Fine-Tuning and Ensembling Strategy for Pancreatic Tumor Segmentation in Diagnostic and Therapeutic MRI}{A Multi-Stage Fine-Tuning and Ensembling Strategy for Pancreatic Tumor Segmentation in Diagnostic and Therapeutic MRI}}
%
%
\author{Omer Faruk Durugol\inst{5,\dagger}\orcidlink{0009-0005-4489-0556}, Maximilian Rokuss\inst{1,2,\dagger,\star}\orcidlink{0009-0004-4560-0760} \and
Yannick Kirchhoff\inst{1,2}\orcidlink{0000-0001-8124-8435} \and\\Klaus H. Maier-Hein\inst{1,3,4}\orcidlink{0000-0002-6626-2463}
}
\authorrunning{O. Durugol et al.}
\titlerunning{DKFZ tackles PANTHER}

\institute{
German Cancer Research Center (DKFZ) Heidelberg,\\Division of Medical Image Computing, Heidelberg, Germany
\and
Faculty of Mathematics and Computer Science,\\Heidelberg University, Heidelberg, Germany\and
Helmholtz Imaging, DKFZ, Heidelberg, Germany\and
Pattern Analysis and Learning Group, Department of Radiation Oncology, Heidelberg University Hospital, Heidelberg, Germany\and International School of Medicine, Istanbul Medipol University, Istanbul, Turkey
}

\maketitle              

\begin{center}
\textsuperscript{\textdagger}These authors contributed equally to this work. \\ 
\textsuperscript{*}Corresponding author: \email{maximilian.rokuss@dkfz-heidelberg.de}

\vspace{1em} 
\end{center}

\begin{abstract}
Automated segmentation of Pancreatic Ductal Adenocarcinoma (PDAC) from MRI is critical for clinical workflows but is hindered by poor tumor-tissue contrast and a scarcity of annotated data. This paper details our submission to the \href{https://panther.grand-challenge.org}{PANTHER challenge}, addressing both diagnostic T1-weighted (Task 1) and therapeutic T2-weighted (Task 2) segmentation. Our approach is built upon the nnU-Net framework and leverages a deep, multi-stage cascaded pre-training strategy, starting from a general anatomical foundation model and sequentially fine-tuning on CT pancreatic lesion datasets and the target MRI modalities. Through extensive five-fold cross-validation, we systematically evaluated data augmentation schemes and training schedules. Our analysis revealed a critical trade-off, where aggressive data augmentation produced the highest volumetric accuracy, while default augmentations yielded superior boundary precision (achieving a state-of-the-art MASD of 5.46 mm and HD95 of 17.33 mm for Task 1). For our final submission, we exploited this finding by constructing custom, heterogeneous ensembles of specialist models, essentially creating a mix of experts. This metric-aware ensembling strategy proved highly effective, achieving a top cross-validation Tumor Dice score of 0.661 for Task 1 and 0.523 for Task 2. Our work presents a robust methodology for developing specialized, high-performance models in the context of limited data and complex medical imaging tasks (\textit{Team MIC-DKFZ}).

\end{abstract}
\renewcommand\thefootnote{\arabic{footnote}}
\setcounter{footnote}{0}
\section{Introduction}
Magnetic Resonance Imaging (MRI) is a cornerstone in the management of pancreatic cancer, one of the most lethal malignancies worldwide\cite{ilic2016epidemiology}. It provides critical visualization for diagnosis, staging, and treatment planning, particularly with the advent of advanced techniques like MR-guided linear accelerators (MR-Linacs) for adaptive radiation therapies\cite{liu2023mri}. However, the manual delineation of tumors and surrounding tissues from these images is a significant bottleneck. This process is not only time-consuming and labor-intensive but also highly dependent on expert radiologist interpretation, which is challenged by the pancreas's complex anatomy and additionally by the subtle appearance of tumors on certain MRI phases\cite{yang2024radpanc}. Automated pancreatic tumor segmentation presents a compelling solution, promising to accelerate clinical workflows and improve the consistency of clinical analysis.\\

\noindent Despite this potential, automated segmentation of pancreatic tumors on MRI faces substantial hurdles. The complexity stems from the low soft-tissue contrast of tumors against surrounding pancreatic tissue, close proximity to similar structures like small/large bowel, anatomical variability between patients, and the significant changes in tumor shape and position that can occur during a course of treatment. Furthermore, the lack of large, publicly available, annotated MRI datasets for pancreatic cancer has severely limited the development, validation, and benchmarking of robust AI-driven solutions\cite{anghel2024data}. Models must be capable of performing accurately across different MRI sequences-such as T1W CE (T1-Weighted, Contrast-enhanced) scans for diagnosis and T2W (T2-Weighted) scans for on-treatment guidance-and generalize well despite variations in imaging phases and protocols.\\

\noindent To spur innovation and address these challenges, the \href{https://panther.grand-challenge.org}{PANTHER challenge} (Pancreatic Tumor Segmentation in Therapeutic and Diagnostic MRI) was established. As the first grand challenge of its kind, \href{https://panther.grand-challenge.org}{PANTHER challenge} provides a platform for researchers to develop and benchmark algorithms on the first publicly available dataset for this task, encompassing both diagnostic T1W and treatment-planning MR-Linac T2W scans. The challenge tasks participants with creating robust models for two distinct clinical needs: Segmenting tumors from T1W diagnostic MRIs (Task 1) and T2W MR-Linac images (Task 2), the latter of which represents a real-world few-shot learning problem due to the limited data available (50 training cases).\\

\noindent This manuscript details our participation in the \href{https://panther.grand-challenge.org}{PANTHER challenge}, where we address both Task 1 and Task 2. Our approach is built upon the strong foundation of the nnU-Net v2 framework. We tackle the aforementioned challenges of pancreatic tumor segmentation through robust data handling, network architecture design, and ensembling strategies. Our goal is to develop highly accurate and generalizable models for both tasks, that can effectively handle the complexities of different MRI sequences and contribute to the advancement of automated tools for pancreatic cancer treatment.

\section{Methods}
\label{sect:methods}
Our method builds upon the robust and self-configuring nnU-Net framework\cite{isensee2021nnu}, a leading approach in medical image segmentation. To address the specific challenges of pancreatic ductal adenocarcinoma (PDAC) segmentation in multi-modal MRI, we developed a multi-stage, cascaded fine-tuning strategy and systematically evaluated the impact of data augmentation and training hyperparameters. We selected the recently introduced Residual Encoder, ResEncL\cite{isensee2024nnu} as our network backbone due to its demonstrated power and large capacity.

\subsection{nnUNet Configuration}
All experiments were conducted using the \verb+3d_fullres+ configuration. A critical step in our methodology was data harmonization; all MRI sequences were resampled to an isotropic spacing of \verb+[1.0, 1.0, 1.0]+ mm. This isotropic resampling ensures that the network learns features that are invariant to the initial orientation and slice thickness of the input scans, which is crucial for building a robust model from multi-center data.\\

\noindent Based on our experimental results, we identified an optimal training schedule of 150-200 epochs with an initial learning rate of 1e-3 and a batch size of 2. We utilized a large, isotropic patch size of \verb+[192, 192, 192]+ to provide the network with sufficient 3D context, which is vital for distinguishing the subtle textural changes of PDAC from surrounding healthy pancreatic tissue. The standard nnU-Net training utilizes a PolyLR scheduler; however, we also explored alternatives with warmup and augmentation to optimize convergence for our specific fine-tuning task.

\subsection{Pretraining and Finetuning}
\label{subsec:pretraining}
\begin{figure}[H] 
    \centering
     \begin{adjustbox}{width=1.02\textwidth, center}
        \includegraphics[width=1.02\textwidth]{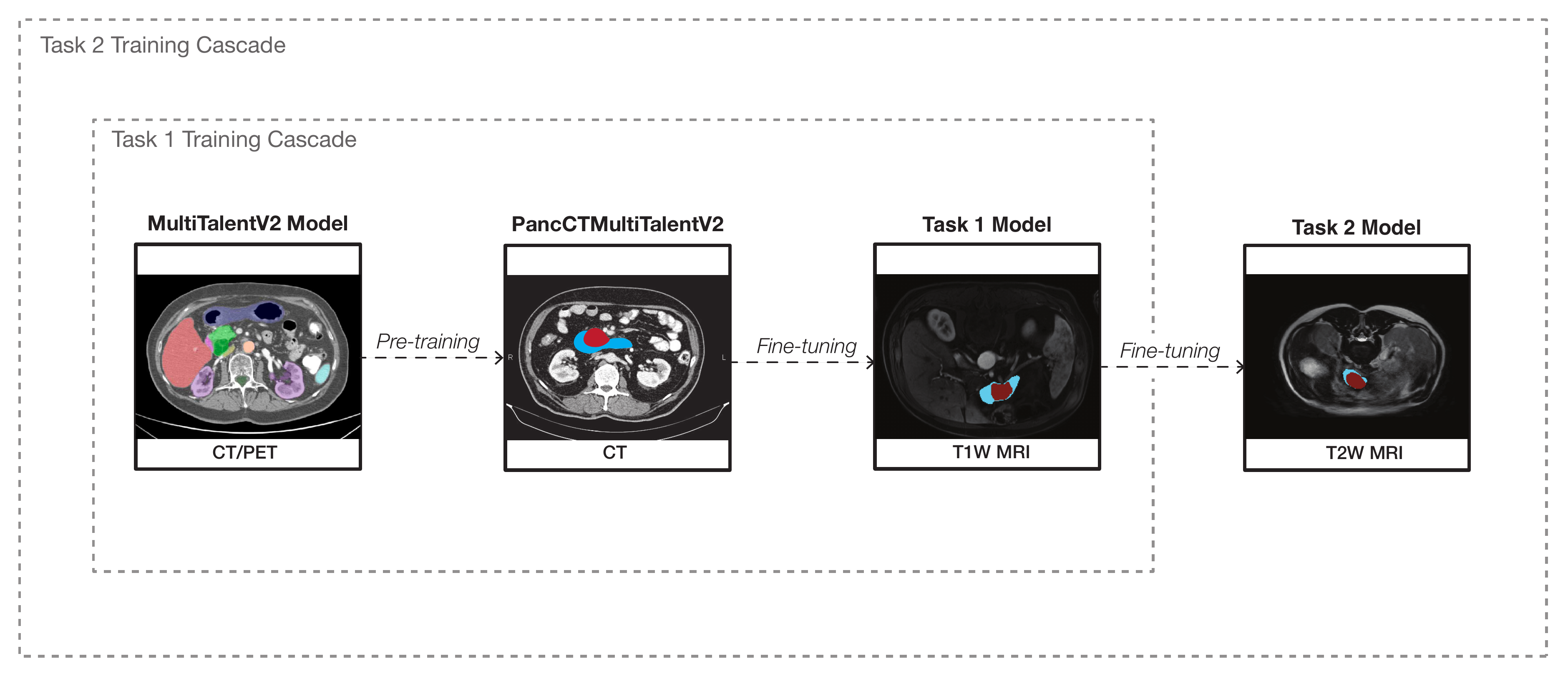}
    \end{adjustbox}
    \caption{
        The multi-stage cascaded fine-tuning strategy for our final models. 
        A foundational model pre-trained on diverse medical data is first specialized on pancreatic CT lesions. 
        This cascade model is then fine-tuned to create the final Task 1 model. 
        Finally, the Task 1 model serves as the foundation for the final Task 2 model, bridging the modality gap from CT to T1W and then to T2W MRI.
    }
    \label{fig:cascade} 
\end{figure}

Recognizing the challenge of segmenting PDAC, especially in the low-contrast T2W modality (Task 2), we adopted a cascaded pre-training and fine-tuning approach to imbue the model with a rich anatomical prior. Our strategy consists of three main stages:

\begin{enumerate}
    \item Foundation Pre-training: We began with a powerful foundation model pre-trained on a vast, multi-modal dataset of over 18,000 3D medical scans, including CT, MRI, and PET images, in a MultiTalent-inspired fashion\cite{ulrich2023}. This model called MultiTalentV2\footnote{\url{https://zenodo.org/records/13753413}}, has a deep, generalized understanding of abdominopelvic anatomy.\\
    \item Task-Specific Pre-training (CT Lesions): The MultiTalentV2 model was then fine-tuned on a large dataset of 760 CT scans combined from MSD and PANORAMA datasets, featuring pancreatic lesions, we call this dataset PancCTPretrain. This step specialized the model's anatomical knowledge to the specific morphology of pancreatic pathologies. The resulting model is our PancCTMultiTalentV2 cascade pre-training model.\\
    \item Modality Bridging and Final Fine-tuning: This PancCTMultiTalentV2 cascade served as the starting point for all our PANTHER-specific training. We reduced the initial learning rate to 1e-3 to counteract catastrophic forgetting and train the model on the learned weights from the pre-trainings. For Task 2 (T2W), we found that an additional intermediate fine-tuning step on Task 1 (T1W) was critical for bridging the modality gap between CT and T2W MRI. Our final and best-performing models for Task1 and Task 2 were therefore trained with the cascades shown in Fig \ref{fig:cascade}:
\end{enumerate}

\noindent We consistently observed that starting training from the \verb+checkpoint_final.pth+ of the bigger foundation models (MultiTalentV2, PancCTMultiTalentV2) performed better for downstream tasks Task 1 \& Task 2, in contrast, Task 2 models had \verb+checkpoint_best.pth+'s outperforming the final checkpoints, likewise, Task 1 had mixed results for best vs. final checkpoints. The best checkpoints performed better especially on Task 2, likely avoiding the propagation of overfitted features, as we observed from the training and validation loss progress.

\subsection{Data Augmentation Ablation Study}\label{subsec:DA}

While the default nnU-Net augmentation scheme provides strong regularization, we hypothesized that the subtle nature of PDAC in T2W MRI might benefit from a more aggressive strategy. We systematically evaluated a suite of heavy data augmentation trainers, referred to as the \verb+DA5+ family. These trainers introduce a higher frequency and intensity of transforms, including blurring, sharpening, and simulated low-resolution artifacts.\\

\noindent We specifically compared three variants which differ in their spatial transform interpolation schemes: \verb+DA5+ (bicubic for image, linear for labels), \verb+DA5ord0+ (nearest-neighbor for both), and \verb+DA5Segord0+ (bicubic for image, nearest-neighbor for labels). This allowed us to assess the trade-off between generating realistic augmented images and preserving precise, artifact-free ground truth boundaries during training.

\subsection{Approaches we tried, that didn't work}
\label{subsec:approaches}

During our extensive experimentation, several promising avenues did not yield a performance benefit for our specific tasks. Fine-tuning a model on both Task 1 and Task 2 data combined, consistently underperformed compared to our sequential, cascaded separate approach. Similarly, for the difficult Task 2, data-balancing strategies such as including all Task 1 data in each training fold of Task 2, or using a 1:1 ratio of Task 1 and Task 2 data in folds of Task 2 training, did not improve performance, suggesting that the quality of the pre-training foundation was more critical than the quantity of out-of-domain training data.\\ 

\noindent We've also tried training only on the tumor labels without using pancreas labels at all. The idea was that the model could perform well in pancreas segmentation, while compromising at the tumor borders instead due to the inherent volumetric relationship of pancreas and PDAC. This performed better for Task 2 at first trials, however as the learning rate, augmentation, and cascade trainings progressed, it started performing worse and was discarded.\\

\noindent In addition, we tried to retrain Task 1 with the unannotated data provided. We used the predicted labels on unannotated data from our best Task 1 model, for 50 of the well labeled cases, which performed worse due to most likely the data including fundamentally different phases than the T1W CE of the Task 1 data. Moreover, since Task 1 unannotated data included different phases during the T1W MRI sequence but from the same patients, we have identified 32 seemingly co-registered images from the unannotated dataset, used ground truth labels as labels for the selected unannotated data, hypothesizing if the tumor boundaries would lie within same area between phases, and retrained; unfortunately, this also performed slightly worse due to the tumor boundaries observed changing from phase to phase, especially considering no contrast and contrast enhanced phases where vascular parts of tumor are observed differently. As last trial of self-annotated training, we tried including pseudolabels only for images the best Task 1 model performed <0.6 tumor Dice score on, this annotated training with 29 images also performed worse than the best model. \\

\noindent Finally, more advanced learning rate schedulers like \verb+CosineAnnealingWithWarmup+ or \verb+PolyLRWithWarmup+ did not outperform the default PolyLR scheduler for this particular fine-tuning problem.

\section{Results}

All models were developed and evaluated using five-fold cross-validation on our nnU-Net assigned splits. The final challenge ranking metrics are Dice Tumor, 5mm Surface Dice, Mean Average Surface Distance (MASD), 95th Percentile Hausdorff Distance (HD95), and Tumor Volume RMSE calculated from \verb+surface-distance+ package as used by challenge organizers. Our analysis focuses on the models that produced the best balance across these five key metrics on the five-fold cross-validation.

\subsection{Isotropic Spacing}

\noindent As shown in Table \ref{tab:isotropic_ablation}, forcing the model to resample all images to a \verb+[1, 1, 1]+ mm spacing and use a large, isotropic \verb+[192, 192, 192]+ patch size yielded a dramatic improvement in performance for Task 1. The mean Tumor Dice score increased by over 7 percentage points, and the boundary-based metrics (Surface Dice, MASD, and HD95) improved substantially, indicating that the model was able to learn more robust and spatially accurate representations.\\

\noindent Interestingly, for the more challenging Task 2, this change had a mixed effect. While the Tumor Dice showed improvement, the boundary metrics declined. This suggests that while isotropic training is clearly superior for the T1W modality, further optimization is required for T2W, which formed the basis of our subsequent experiments.

\begin{table}[t]
    \begin{center}
    \caption{Impact of isotropic resampling on model performance. 
        This table compares the five-fold cross-validation results of our baseline model trained with the default anisotropic spacing versus the same model trained with a forced isotropic spacing of \texttt{[1.0, 1.0, 1.0]} mm and a corresponding \texttt{[192, 192, 192]} patch size. 
        All metrics are averaged across the five folds. Arrows indicate the desired direction for each metric. Best results are in \textbf{bold}.}
    \label{tab:isotropic_ablation}
    \adjustbox{max width=\textwidth}{%
        \begin{tabular}{l|ccccc|ccccc}
            \hline
             & \multicolumn{5}{c|}{\textbf{Task 1 (T1W MRI)}} & \multicolumn{5}{c}{\textbf{Task 2 (T2W MRI)}} \\
            \textbf{Setting} & Dice$\uparrow$ & Surf Dice$\uparrow$ & MASD$\downarrow$ & HD95$\downarrow$ & RMSE$\downarrow$ & Dice$\uparrow$ & Surf Dice$\uparrow$ & MASD$\downarrow$ & HD95$\downarrow$ & RMSE$\downarrow$ \\
            \hline
            Default & 0.495 & 0.679 & 117.58 & 132.40 & \textbf{28905} & 0.339 & \textbf{0.536} & \textbf{66.98} & \textbf{82.22} & 20177 \\
            Isotropic & \textbf{0.570} & \textbf{0.770} & \textbf{49.41} & \textbf{64.32} & 29385 & \textbf{0.352} & 0.516 & 126.33 & 140.76 & \textbf{20174} \\
            \hline
        \end{tabular}
    }
    \end{center}
\end{table}

\subsection{Learning Rate}

\noindent A critical hyperparameter in any fine-tuning procedure is the initial learning rate. While the default nnU-Net learning rate of 1e-2 is optimized for training from scratch, we hypothesized that a lower rate would be necessary to carefully adapt our powerful pre-trained models without causing instability or catastrophic forgetting. We therefore conducted an ablation study on fine-tuning Task 1 and 2, comparing the default rate with a reduced rate of 1e-3.\\

\noindent As presented in Table \ref{tab:lr_ablation}, the results of this experiment were unambiguous. For Task 1, reducing the learning rate led to a substantial improvement across all five key metrics, increasing the Tumor Dice by over 5 percentage points and dramatically reducing both boundary errors and the Tumor RMSE. The positive effect was even more pronounced for the more challenging Task 2, where the MASD and HD95 were nearly halved, indicating a much more stable and effective convergence. Based on this clear evidence, we adopted 1e-3 as the standard initial learning rate for all subsequent fine-tuning experiments in our study.

\begin{table}[t]
    \centering
    \caption{Ablation study on the initial learning rate. 
        We compare the performance of fine-tuning using the nnU-Net default learning rate of \texttt{1e-2} against a reduced rate of \texttt{1e-3}. 
        This experiment was conducted on a baseline model pre-trained on the MultiTalentV1/V2 datasets for Task1 \& Task2 respectively. 
        The results show a clear and substantial benefit to using a lower learning rate for this fine-tuning task.}
    \label{tab:lr_ablation}
    \adjustbox{max width=\textwidth}{%
        \begin{tabular}{l|ccccc|ccccc}
            \hline
             & \multicolumn{5}{c|}{\textbf{Task 1 (T1W MRI)}} & \multicolumn{5}{c}{\textbf{Task 2 (T2W MRI)}} \\
            \textbf{Setting} & Dice$\uparrow$ & Surf Dice$\uparrow$ & MASD$\downarrow$ & HD95$\downarrow$ & RMSE$\downarrow$ & Dice$\uparrow$ & Surf Dice$\uparrow$ & MASD$\downarrow$ & HD95$\downarrow$ & RMSE$\downarrow$ \\
            \hline
            LR (\texttt{1e-2}) & 0.570 & 0.770 & 49.41 & 64.32 & 29385 & 0.368 & 0.539 & 87.66 & 102.07 & 19764 \\
            LR (\texttt{1e-3}) & \textbf{0.622} & \textbf{0.811} & \textbf{36.32} & \textbf{48.14} & \textbf{18439} & \textbf{0.403} & \textbf{0.589} & \textbf{45.93} & \textbf{60.19} & \textbf{18595} \\
            \hline
        \end{tabular}
    }
\end{table}

\subsection{Training for Task 1:}
\noindent Our final model development for Task 1 revealed a distinct trade-off between volumetric and boundary-based performance, contingent on the data augmentation strategy and training duration. As presented in Table \ref{tab:task1_results}, different configurations excelled in different aspects of the segmentation challenge.\\

\noindent The model trained for 1000 epochs with the heavily regularizing DA5ord0 augmentation scheme achieved unparalleled boundary precision, yielding the best Mean Average Surface Distance (MASD) and 95th Percentile Hausdorff Distance (HD95) by a substantial margin. Conversely, a shorter 200-epoch run using the standard DA5 augmentation on our Pancbest (PancCTMultiTalentV2 cascade best checkpoint instead of final) foundation model produced the highest volumetric overlap, achieving the best Tumor Dice and Surface Dice scores.\\

\noindent Ultimately, we selected the best checkpoints of the model trained for 1000 epochs with default nnU-Net augmentations as our final submission model for Task 1. While not the absolute best in any single category, it demonstrated the most robust and balanced performance across all five key metrics. It achieved an excellent Tumor Dice of 0.656 and Surface Dice of 0.841, produced the lowest Tumor RMSE of 14850, and maintained elite boundary metrics with a MASD of 15.19 mm and an HD95 of 25.77 mm. The extended training duration, combined with the proven stability of the default augmentation scheme, allowed this model to converge to a highly accurate and generalizable state, making it our most reliable all-around performer.

\begin{table}[t]
    \centering
    \caption{
        Final five-fold cross-validation results for our top-performing Task 1 models. 
        The table highlights the trade-off between models optimized for volumetric accuracy (Dice Tumor, RMSE) versus those optimized for boundary precision (MASD, HD95). \textit{"Pancbest":} Model trained on PancCTMV2 cascade model's best checkpoint instead of final. \textit{"best.pth":} indicates best checkpoint performance instead of default "final.pth".
        All metrics are averaged across the five folds. Arrows indicate the desired direction. Best performance for each metric is in \textbf{bold}.
    }
    \label{tab:task1_results}
    \adjustbox{max width=\textwidth}{%
        \begin{tabular}{l|ccccc}
            \hline
            \textbf{Setting} & Dice Tumor$\uparrow$ & Surf Dice$\uparrow$ & MASD$\downarrow$ & HD95$\downarrow$ & RMSE$\downarrow$ \\
            \hline
            Default @ 1000 epochs & 0.628 & 0.822 & 15.49 & 26.32 & 19334 \\
            Default @ 1000 epochs (best.pth) & 0.656 & 0.841 & 15.19 & 25.77 & \textbf{14850} \\
            DA5 @ 200 epochs (Pancbest, best.pth) & \textbf{0.661} & \textbf{0.843} & 15.35 & 27.92 & 19330 \\
            DA5ord0 @ 1000 epochs (best.pth) & 0.651 & 0.832 & \textbf{5.46} & \textbf{17.33} & 20785 \\
            \hline
        \end{tabular}
    }
\end{table}

\begin{table}[t]
    \centering
    \caption{
        Five-fold cross-validation results for our top-performing Task 2 models. 
        The results show that fine-tuning on a strong Task 1 model with heavy data augmentation (`DA5`) was the most effective strategy. 
        Our final model strikes a superior balance between volumetric and boundary-based performance.\\
        \textit{"best.pth":} indicates best checkpoint performance instead of default "final.pth".
        Metrics are averaged across five folds. Arrows indicate the desired direction. Best performance is in \textbf{bold}.
    }
    \label{tab:task2_results}
    \adjustbox{max width=\textwidth}{%
        \begin{tabular}{l|ccccc}
            \hline
            \textbf{Setting (Fine-tuned from Task 1)} & Dice Tumor$\uparrow$ & Surf Dice$\uparrow$ & MASD$\downarrow$ & HD95$\downarrow$ & RMSE$\downarrow$ \\
            \hline
            Default @ 1000 epochs & 0.429 & 0.615 & 26.04 & 41.06 & 18936 \\
            Default @ 150 epochs (best.pth) & 0.491 & 0.682 & 4.74 & \textbf{17.27} & 17905 \\
            DA5 @ 150 epochs (best.pth)\textsuperscript{*} & 0.501 & 0.697 & \textbf{4.63} & 18.93 & 16899 \\
            DA5Segord0 @ 150 epochs (best.pth)& \textbf{0.523} & \textbf{0.698} & 24.60 & 38.85 & \textbf{15415} \\
            \hline
        \end{tabular}
    }
    \begin{threeparttable}
   {\scriptsize \textsuperscript{*}This model was fine-tuned on a different Task 1 foundation model (\verb+Task1-200e_DA5_best+) compared to the others, which were based on the (\verb+Task1-1000e_best+) model.}
   \end{threeparttable}
\end{table}

\subsection{Training for Task 2:}

\noindent The segmentation of PDAC in T2W MRI proved to be a significantly more challenging problem, revealing a distinct trade-off in our models between volumetric accuracy and boundary precision. Our experiments consistently demonstrated that a further cascaded approach, fine-tuning on a well-rounded model already optimized for Task 1, was essential for success.\\

\noindent Our analysis, summarized in Table \ref{tab:task2_results}, showed that a model trained for 150 epochs with the aggressive DA5Segord0 data augmentation was unequivocally superior on the volume- and overlap-centric metrics that are critical for assessing tumor burden. This configuration produced our final selected model for the Task 2 submission.\\

\noindent This champion model achieved a vast Dice Tumor score of 0.523 and the highest Surface Dice of 0.698 among all our experiments. Furthermore, it attained the lowest (best) Tumor RMSE of 15415, indicating a superior ability to estimate the overall tumor volume. This state-of-the-art volumetric performance, however, came at the cost of boundary precision, resulting in higher MASD (24.60 mm) and HD95 (38.85 mm) values compared to models trained with less aggressive augmentation.\\

\noindent Ultimately, we selected this model for our final submission due to its dominant performance on metrics related to volumetric and overlap accuracy. We prioritized the substantial gains in Dice Tumor and Tumor RMSE, accepting the trade-off of lower precision on the fine-grained boundary distance metrics.

\subsection{Test Set Submission}

\noindent For the final test set submission, we constructed custom, high-performance heterogeneous ensembles for each task, designed to maximize performance across the full spectrum of evaluation metrics. This approach was carefully optimized to ensure inference for each case was completed within the 5-minute time constraint, and as such, no test-time augmentation such as mirroring was utilized.\\

\noindent For \textbf{Task 1}, we created a specialized 6-fold ensemble by selecting the best individual cross-validation folds from our top-performing models, each of which excelled at a different aspect of the segmentation challenge. This "expert" ensemble was composed of:
\begin{itemize}
    \item Three folds from our best "All-Rounder" model (\verb+1000e_best+), which demonstrated the most balanced overall performance.
    \item Two folds from our "Volume” model (\verb+Pancbest_DA5_200e_best+), which was superior at capturing the full tumor volume.
    \item One fold from our "Boundary" model (\verb+DA5ord0_1000e_best+), which produced the most precise and anatomically plausible boundaries.
\end{itemize}

\noindent For \textbf{Task 2}, we employed a similar strategy of constructing a custom heterogeneous ensemble. Folds were selected from our top-performing models, including:
\begin{itemize}
    \item One fold from best "All-Rounder" model (\verb+Task1_DA5_200e-DA5_150e_best+), which demonstrated the most balanced overall performance.
    \item Four folds from our "Volume” model (\verb+Task1_1000e-DA5Segord0_150e_best+), which was superior at capturing the full tumor volume.
\end{itemize}

\noindent The final composition was chosen to create the optimal balance between high volumetric accuracy (Dice Tumor, RMSE) and elite boundary precision (Surface Dice, MASD, HD95), resulting in a robust and highly competitive model for the final evaluation.

\section{Conclusion}

\noindent In this paper, we presented a robust methodology for the segmentation of pancreatic ductal adenocarcinoma in multi-modal MRI, developed for our submission to the \href{https://panther.grand-challenge.org}{PANTHER challenge}. Our approach was built upon a rigorous, multi-stage fine-tuning strategy within the nnU-Net framework, combined with a systematic evaluation of data augmentation and training hyperparameters.\\

\noindent Our key findings demonstrate that a deep, cascaded pre-training approach, beginning with a general anatomical foundation model and sequentially fine-tuning on task- and modality-specific datasets, was essential for achieving state-of-the-art performance. We discovered a critical trade-off between aggressive data augmentation, which excelled at capturing tumor volume, and default augmentation, which produced superior boundary precision. By carefully analyzing the five key challenge metrics, we were able to select and combine the strengths of our best models into powerful, heterogeneous ensembles for the final submission.\\

\noindent This holistic approach, which balanced a sophisticated pre-training cascade with targeted, metric-driven ensembling, resulted in a highly competitive and robust solution for both T1-weighted and the more challenging T2-weighted segmentation tasks. Our work underscores the importance of cascaded transfer learning and demonstrates the power of metric-aware ensembling in advancing the state-of-the-art in medical image segmentation.
    
\begin{credits}
\subsubsection{\ackname}
This work was conducted during a research internship at the Division of Medical Image Computing at the German Cancer Research Center (DKFZ), Heidelberg. Part of this work was funded by Helmholtz Imaging (HI), a platform of the Helmholtz Incubator on Information and Data Science. The present contribution is supported by the Helmholtz Association under the joint research school "HIDSS4Health – Helmholtz Information and Data Science School for Health".
\end{credits}

\newpage
\bibliographystyle{splncs04}
\bibliography{biblio}

\begin{thebibliography}{1}
\providecommand{\url}[1]{\texttt{#1}}
\providecommand{\urlprefix}{URL }
\providecommand{\doi}[1]{https://doi.org/#1}

\bibitem{anghel2024data}
Anghel, C., Grasu, M.C., Anghel, D.A., Rusu-Munteanu, G.I., Dumitru, R.L., Lupescu, I.G.: Pancreatic adenocarcinoma: Imaging modalities and the role of artificial intelligence in analyzing ct and mri images. Diagnostics  \textbf{14}(4) (2024). \doi{10.3390/diagnostics14040438}, \url{https://www.mdpi.com/2075-4418/14/4/438}

\bibitem{ilic2016epidemiology}
Ilic, M., Ilic, I.: Epidemiology of pancreatic cancer. World journal of gastroenterology  \textbf{22}(44), ~9694 (2016)

\bibitem{isensee2021nnu}
Isensee, F., Jaeger, P.F., Kohl, S.A.A., Petersen, J., Maier-Hein, K.H.: {nnU-Net}: a self-configuring method for deep learning-based biomedical image segmentation. Nat. Methods  \textbf{18}(2),  203--211 (Feb 2021)

\bibitem{isensee2024nnu}
Isensee, F., Wald, T., Ulrich, C., Baumgartner, M., Roy, S., Maier-Hein, K., Jäger, P.F.: { nnU-Net Revisited: A Call for Rigorous Validation in 3D Medical Image Segmentation }. In: proceedings of Medical Image Computing and Computer Assisted Intervention -- MICCAI 2024. vol. LNCS 15009. Springer Nature Switzerland (October 2024)

\bibitem{liu2023mri}
Liu, X., Li, Z., Yin, Y.: Clinical application of {MR-Linac} in tumor radiotherapy: a systematic review. Radiat. Oncol.  \textbf{18}(1), ~52 (Mar 2023)

\bibitem{ulrich2023}
Ulrich, C., Isensee, F., Wald, T., Zenk, M., Baumgartner, M., Maier-Hein, K.H.: MultiTalent: A Multi-dataset Approach to Medical Image Segmentation, p. 648–658. Springer Nature Switzerland (2023). \doi{10.1007/978-3-031-43898-1_62}

\bibitem{yang2024radpanc}
Yang, D.H., Park, S.H., Yoon, S.: Differential diagnosis of pancreatic cancer and its mimicking lesions. J Korean Soc Radiol  \textbf{85}(5),  902--915 (Sep 2024)

\end{thebibliography}

\end{document}